\documentclass[11pt,a4paper]{article}
\usepackage{xr-hyper}
\usepackage[hyperref]{acl2021}
\usepackage{times}
\usepackage{latexsym}
\usepackage{bbding,pifont,adjustbox,amssymb} 
\usepackage{url}

\usepackage{microtype}

\usepackage[T1]{fontenc}
\usepackage[scaled=.8]{beramono}
\usepackage[scaled=.95]{helvet}
\usepackage{amsmath}
\usepackage{color,soul}

\usepackage{subcaption}

\usepackage[shortlabels]{enumitem} 
\setitemize{noitemsep,topsep=0em} 

\usepackage{multirow}
\usepackage{tikz}
\usepackage{tikz-dependency}
\usepackage[warn]{textcomp}
\usepackage{etoolbox}
\usepackage{xfrac}
\usepackage{pgfplotstable}

\usepackage{tabularx}
\usepackage{collcell}
\usepackage{booktabs}
\usepackage{nameref}
\usepackage{cleveref}

\pgfplotsset{compat=1.14}

\usepackage{wrapfig}

\makeatletter
\newcommand*{\addFileDependency}[1]{
  \typeout{(#1)}
  \@addtofilelist{#1}
  \IfFileExists{#1}{}{\typeout{No file #1.}}
}
\makeatother

\aclfinalcopy 

\title{Great Service! Fine-grained Parsing of Implicit Arguments}

\author{
  Ruixiang Cui \and Daniel Hershcovich \\
  Department of Computer Science \\
  University of Copenhagen \\
  \quad\texttt{\{rc, dh\}@di.ku.dk} \\
}

\date{}

\begin{document}
\maketitle
\begin{abstract}
Broad-coverage meaning representations in NLP mostly focus on explicitly expressed content. More importantly, the scarcity of datasets annotating diverse implicit roles limits empirical studies into their linguistic nuances.  For example, in the web review ``Great service!'', the provider and consumer are implicit arguments of different types. We examine an annotated corpus of fine-grained implicit arguments \cite{cui-hershcovich-2020-refining} by carefully re-annotating it, resolving several inconsistencies. Subsequently, we present the first transition-based neural parser that can handle implicit arguments dynamically, and experiment with two different transition systems on the improved dataset. We find that certain types of implicit arguments are more difficult to parse than others and that the simpler system is more accurate in recovering implicit arguments, despite having a lower overall parsing score, attesting current reasoning limitations of NLP models. This work will facilitate a better understanding of implicit and underspecified language, by incorporating it holistically into meaning representations.

\end{abstract}

\section{Introduction}\label{sec:intro}

Studies of form and meaning, the dual perspectives of the language sign, can be traced back to modern linguistics since \citet{de1916course,de1978course}. They are relevant to modern NLP, as even current large neural language models cannot intrinsically achieve a human-analogous understanding of natural language \citep{vzabokrtsky2020sentence,bender2020climbing}.
Computational linguists attempt to capture syntactic and semantic features by means of constructing meaning representation frameworks, such as PTG \citep{bohmova2003prague}, EDS \citep{oepen-lonning-2006-discriminant}, AMR \citep{banarescu2013abstract} and UCCA \citep{abend2013universal}.
Through such frameworks, researchers have been exploring linguistic phenomena such as quantification \citep{pustejovsky-etal-2019-modeling}, coreference \citep{prange-etal-2019-semantically}, and word sense  \citep{schneider-etal-2018-comprehensive}. 

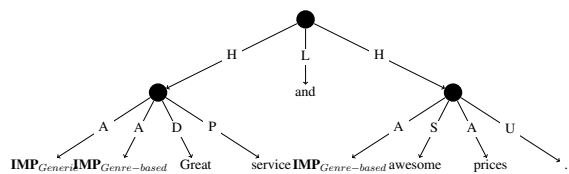
\begin{figure}
\resizebox{7.5cm}{!}{%
\begin{tikzpicture}[->,level distance=2cm,
  level 1/.style={sibling distance=4cm},
  level 2/.style={sibling distance=20.5mm},
  level 3/.style={sibling distance=21mm},
  every circle node/.append style={fill=black},
  every node/.append style={text height=1ex,text depth=0}]
  \tikzstyle{word} = [font=\rmfamily,color=black]
  \node (1_1) [circle] {}
  {
  child {node (1_2) [circle] {}
    {
    child {node (1_8) [word] {\textbf{IMP$_{Generic}$}}  edge from parent node[midway, fill=white]  {A}}
    child {node (1_9) [word] {\textbf{IMP$_{Genre-based}$}}  edge from parent node[midway, fill=white]  {A}}
    child {node (1_10) [word] {Great}  edge from parent node[midway, fill=white]  {D}}
    child {node (1_11) [word] {service}  edge from parent node[midway, fill=white]  {P}}
    } edge from parent node[midway, fill=white]  {H}}
  child {node (1_3) [word] {and}  edge from parent node[midway, fill=white]  {L}}
  child {node (1_4) [circle] {}
    {
    child {node (1_5) [word] {\textbf{IMP$_{Genre-based}$}}  edge from parent node[midway, fill=white]  {A}}
    child {node (1_6) [word] {awesome}  edge from parent node[midway, fill=white]  {S}}
    child {node (1_7) [word] {prices}  edge from parent node[midway, fill=white]  {A}}
    child {node (1_12) [word] {.}  edge from parent node[midway, fill=white]  {U}}
    } edge from parent node[midway, fill=white]  {H}}
  };
\end{tikzpicture}
}
\caption{Example of a UCCA graph with fine-grained implicit arguments.}
    \label{fig:example 1}
\end{figure}

\begin{table}[t]
\centering
\scalebox{0.7}{
\begin{tabular}{ll|ll}
Participant & A & Linker & L \\
Center & C & Connector & N \\
Adverbial & D & Process & P \\
Elaborator & E & Quantifier & Q \\
Function & F & Relator & R \\
Ground & G & State & S \\
Parellel Scene & H & Time & T\\ \midrule
\multicolumn{2}{l|}{Deictic} & \multicolumn{2}{l}{Generic} \\ \multicolumn{2}{l|}{Genre-based} & \multicolumn{2}{l}{Type-identifiable} \\ \multicolumn{2}{l|}{Non-specific} & \multicolumn{2}{l}{Iterated-set} \\
\end{tabular}
}
\caption{UCCA Foundational Layer categories (above) and Implicit Participant Refinement Layer categories (below).}
\label{tab:categories}
\end{table}
In recent years, a few studies have been dedicated to annotating and modelling implicit and underspecified language \citep{roesiger-etal-2018-bridging,elazar-goldberg-2019-wheres,mcmahan-stone-2020-analyzing}. For example, in the online review ``there is no delivery'', two of the omitted elements are the business agent and delivered items. However, previous works focus on specific phenomena requiring linguistic and collaborative reasoning, e.g., bridging resolution, numeric fused-heads identification and referential communication. Such datasets overexpose models to limited linguistic expressions, without leveraging the complete syntactic and semantic features of the context, regardless of the diversity of implicit roles. 

\citet{o2019bringing} and \citet{cui-hershcovich-2020-refining} present works on fine-grained implicit role typology, incorporating it into the meaning representation frameworks AMR and UCCA, respectively. They lay a foundation for the interpretation of idiosyncratic behaviours of implicit arguments from a linguistic and cognitive perspective. Nevertheless, neither provided a dataset ready for computational studies of such implicit arguments.

We take the latter as a starting point, addressing several theoretical inconsistencies, and evaluate its applicability by carefully re-annotating their pilot dataset and providing inter-annotator agreement. Our categorisation set, consisting of six implicit role types, is compatible with UCCA's semantic notion of Scene rather than specific linguistic phenomena. Furthermore, as opposed to previous work, it tackles only essential implicit arguments, salient in cognitive processing.

We design the first semantic parser, with two different transition systems, that has the ability to parse fine-grained implicit arguments for meaning representations and evaluate its performance on the revisited dataset. To conclude, we reflect on this work's objectives and the challenges to face in future research on implicit arguments\footnote{Revisited Implicit EWT dataset is available on \url{https://github.com/ruixiangcui/UCCA-Refined-Implicit-EWT_English}. The code for the implicit parser is on \url{https://github.com/ruixiangcui/implicit_parser}}.

\section{Revisiting Implicit Argument Refinement}\label{sec:refinement}

Universal Conceptual Cognitive Annotation \citep[UCCA;][]{abend2013universal}, a typologically-motivated meaning representation framework, has been targeted by several parsing shared tasks \cite{hershcovich-etal-2019-semeval,oepen-etal-2019-mrp,oepen-etal-2020-mrp}.
It uses directed acyclic graphs (DAG) anchored in surface tokens,
where labelled edges represent semantic relations. In the \textit{foundational layer}, these are based on the notion of Scenes (States and Processes), their Participants and modifiers.
UCCA distinguishes \textit{primary} edges, corresponding to explicit content, from \textit{remote} edges, allowing cross-Scene relations.
Additionally, \textit{Implicit} units represent entities of importance to the interpretation of a Scene that are not explicitly anchored in the text.
Several refinement layers have been proposed beyond the foundational layer, adding distinctions important for semantic roles and coreference \cite{shalev-etal-2019-preparing,prange-etal-2019-semantically}.

\subsection{Fine-grained Implicit Argument Refinement for UCCA}
\citet{cui-hershcovich-2020-refining} proposed a fine-grained implicit argument typology implemented as a refinement layer of Participants for UCCA, centering around the semantic notion of Scene.
Their proposed categorisation set 
consisting of six types, listed in Table~\ref{tab:categories}, is argued to have low annotation complexity and ambiguity, thus requiring relatively low cognitive load for annotation than other fine-grained implicit argument typologies \citep{o2019bringing}.

For example, the online review ``Great service and awesome price!'' is annotated as follows and visualised as Figure~\ref{fig:example 1}:
\begin{enumerate}
    \item[(1)] {[}Great$_D$ service$_P$ \textit{IMP$_{Generic}$} \textit{IMP$_{Genre-based}$}]$_H$ and$_L$ [awesome$_S$ prices$_A$ \textit{IMP$_{Genre-based}$}]$_H$! 
\end{enumerate}
There are two Scenes invoked in the sentence. ``Great service'' with ``service'' as a Process, and ``awesome prices'' with ``awesome'' as a State. Who is serving, who is being served and what is priced are distinct implicit arguments (\textit{IMP}), which require reasoning to resolve. 

\textit{Genre-based} roles refer to conventional omission in the genre \citep{ruppenhofer2010constructional}. The corpus is based on online reviews, where reviewers typically do not mention what is under review. Therefore, an implicit argument in each Scene is marked as Genre-based referring to the reviewee. \textit{Generic} roles denote ``people in general'' \citep{lambrecht2005definite}. In the example, the recipient could be anyone, rather than a specific person.

\subsection{Revisiting Inconsistencies}

Despite the general soundness of \citet{cui-hershcovich-2020-refining}'s typology, we find multiple inconsistencies in it: prominently, the treatment of Process and State Scenes and of nominalisation, and some borderline cases. We propose to revisit these cases and introduce consistent implicit argument refinement guidelines.

\subsubsection{State Scenes and Process Scenes} 
UCCA differentiates Scenes according to their persistency in time \cite{abend2013universal}: stative Scenes are temporally persistent states, while processual Scenes are evolving events. \citet{cui-hershcovich-2020-refining} did not annotate implicit arguments in State Scenes, although they are essentially similar. We, therefore, categorise them using the same guidelines. This phenomenon is rather pervasive in the dataset, as in Example 1 and 2:
\begin{enumerate}
    \item[(2)] Very$_D$ friendly$_S$ [[even$_E$ at$_C$]$_R$ weekends$_C$]$_T$ \textit{IMP$_{Genre-based}$}.
\end{enumerate}

\subsubsection{The Definition of Prominent Elements}
While UCCA only annotates implicit units when they are ``prominent in the interpretation of the Scene'' \cite{abend2013universal}, it is not always clear what should be regarded as such: the level of uniqueness plays an essential role in recognising the prominence of an argument. For example, even distinguishing definite and indefinite articles may pose a challenge to semantic analysis \citep{SeeminglyIndefiniteDefinites}. There are some linguistic tests for testing whether an argument is semantically mandatory \citep{goldberg2001patient, hajic-etal-2012-announcing}. As for UCCA, we posit that time, location and instrument modifiers, as a rule, are ubiquitous to such an extent that no implicit argument should be annotated unless a Process or State warrants it. In Example 3 (visualised as the upper graph in Figure~\ref{fig:Evaluation example 1}), in the Scene ``you leave,'' the departing action demands that the implicit source location is vital for understanding, unlike the location element in Example 1 or 2, where they are of low prominence.
\begin{enumerate}
    \item[(3)] [(\textit{You})$_A$ have$_D$ [[a$_F$... mechanic$_C$] real$_D$]$_{P/A}$ check$_P$ \textit{IMP}$_{Non-specific}$]$_H$  before$_L$ [you$_A$ leave$_P$ \textit{IMP}$_{Non-specific}$]$_H$.
\end{enumerate}

\subsubsection{Nominalisation as Agent Nouns}
An Agent noun derives from another word denoting an action, and that identifies an entity that perform that action. In UCCA, such an expression is annotated as a single unit with both the Process and Participant categories. As in Example 3, ``a real mechanic'' is marked P/A.
Like time or location, it is subjective whether and how many implicit arguments should be annotated given the possible list of  involved receivers, instruments, etc.

For a profession like ``mechanic'', whatever is being repaired, as well as the repair tools, can be omitted. While an agent noun can invoke a Scene, they could also simply serve as a title. Such cases, e.g. ``doctor,'' ``professor'' and ``chairman,'' are common. To facilitate consistent annotation, we decide never to annotate implicit Participants in Scenes licensed by agent nouns. This means that there is just one Participant, the person themselves.

\subsubsection{Indefinite Deictic Participant}
Deictic arguments refer to the implicit speaker or addressee. They may be confused with Generic roles when it is ambiguous whether the speaker or addressee participates in the Scene. As a rule, we posit that an implicit Participant ought to be Generic unless it evidently refers to the speaker or addressee. Example 4 shows an ambiguous case.
\begin{enumerate}
    \item[(4)] [The$_F$ experience$_C$]$_P$ \textit{IMP}$_{Generic}$ [with$_R$ every$_Q$ department$_C$]$_A$ has$_F$ been$_F$ great$_D$. 
\end{enumerate}

\section{Fine-grained Implicit Argument Corpus}

As a pilot annotation experiment,
\citet{cui-hershcovich-2020-refining} reviewed and refined 116 randomly selected passages from an UCCA-annotated dataset of web reviews \cite[UCCA EWT;][]{hershcovich2019content}.
The dataset was annotated by only one annotator.
With our revisited guidelines, we ask two annotators\footnote{The annotators have a background in linguistics and cognitive science. One has experience in annotating UCCA's Foundational Layer. Training for the other took \textasciitilde10 hours.} to revisit the original dataset, adding or modifying implicit arguments, and subsequently refining their categories, using UCCAApp \citep{abend-etal-2017-uccaapp}.

\subsection{Evaluation of Implicit Argument Annotation}
\label{sec:Evaluation}

Standard UCCA evaluation compares two graphs (e.g., created by different annotators, or one being the gold annotation and the other predicted by a parser), providing an F1 score by matching the edges by their terminal span and category \cite{hershcovich-etal-2019-semeval}.
However, the standard evaluation completely ignores implicit argument annotation. To quantify inter-annotator agreement and later parser performance (\S\ref{sec:results}), we provide an evaluation metric taking these units into account.

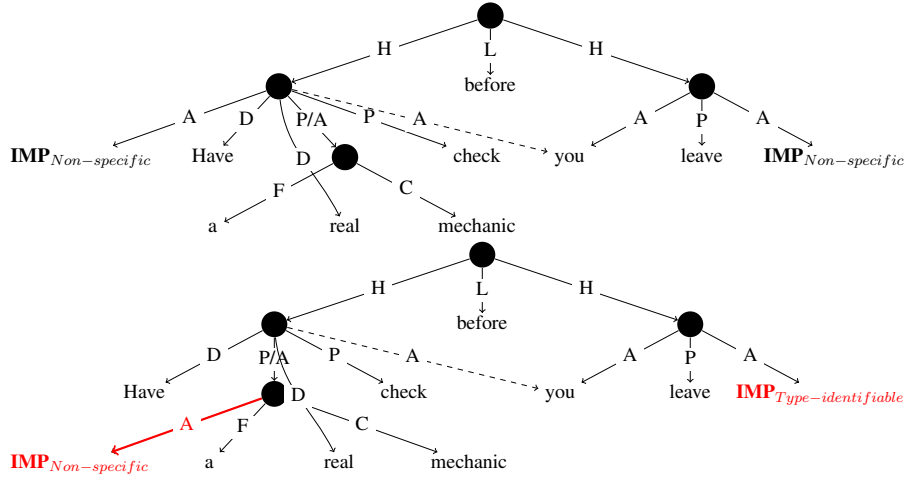
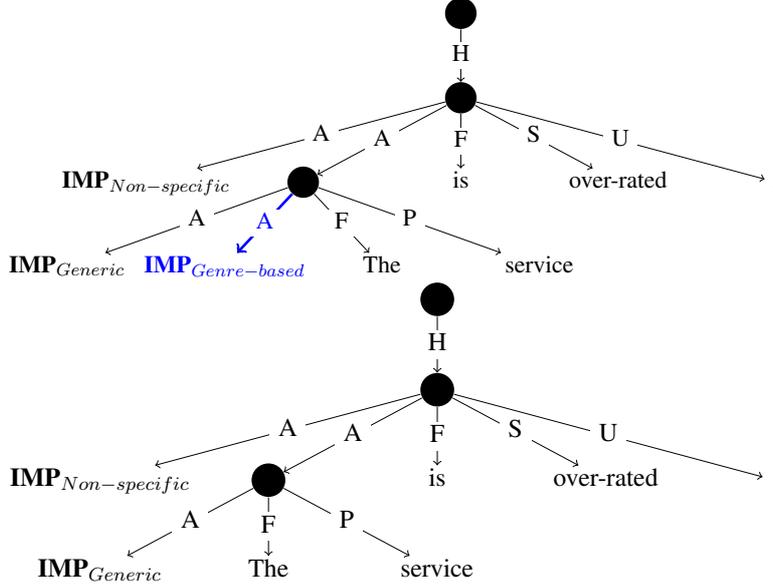
\begin{figure*}[!t]
\centering
\begin{subfigure}[b]{0.8\textwidth}
\resizebox{12cm}{!}{%
\begin{tikzpicture}[->,level distance=1.35cm,
  level 1/.style={sibling distance=4cm},
  level 2/.style={sibling distance=25mm},
  level 3/.style={sibling distance=25mm},
  every circle node/.append style={fill=black},
  every node/.append style={text height=1ex,text depth=0}]
  \tikzstyle{word} = [font=\rmfamily,color=black]
  \node (1_1) [circle] {}
  {
  child {node (1_2) [circle] {}
    {
    child {node (1_16) [word] {\textbf{IMP$_{Non-specific}$}}  edge from parent node[midway, fill=white]  {A}}
    child {node (1_8) [word] {Have}  edge from parent node[midway, fill=white]  {D}}
    child {node (1_9) [circle] {}
      {
      child {node (1_13) [word] {a}  edge from parent node[midway, fill=white]  {F}}
      child {node (1_12) [word] {real}  edge from parent [white]}
      child {node (1_14) [word] {mechanic}  edge from parent node[midway, fill=white]  {C}}
      } edge from parent node[midway, fill=white]  {P/A}}
    child {node (1_10) [word] {check}  edge from parent node[midway, fill=white] {P}}
    } edge from parent node[midway, fill=white]  {H}}
  child {node (1_3) [word] {before}  edge from parent node[midway, fill=white]  {L}}
  child {node (1_4) [circle] {}
    {
    child {node (1_6) [word] {you}  edge from parent node[midway, fill=white]  {A}}
    child {node (1_7) [word] {leave}  edge from parent node[midway, fill=white] {P}}
    child {node (1_17) [word] {\textbf{IMP$_{Non-specific}$}}  edge from parent node[midway, fill=white]  {A}}
    } edge from parent node[midway, fill=white]  {H}}
  };
  \draw[dashed,->] (1_2) to node [midway, fill=white] {A} (1_6);
  \draw[bend right,->] (1_2) to[out=-20, in=180] node [midway, fill=white] {D} (1_12);
\end{tikzpicture}
}
\resizebox{12cm}{!}{%
\begin{tikzpicture}[->,level distance=1.35cm,
  level 1/.style={sibling distance=4cm},
  level 2/.style={sibling distance=25mm},
  level 3/.style={sibling distance=25mm},
  every circle node/.append style={fill=black},
  every node/.append style={text height=1ex,text depth=0}]
  \tikzstyle{word} = [font=\rmfamily,color=black]
  \node (1_1) [circle] {}
  {
  child {node (1_2) [circle] {}
    {
    child {node (1_8) [word] {Have}  edge from parent node[midway, fill=white]  {D}}
    child {node (1_9) [circle] {}
      {
      child {node (1_16) [word][text=red] {\textbf{IMP$_{Non-specific}$}}  edge from parent node[midway, fill=white]  {A}}
      child {node (1_13) [word] {a}  edge from parent node[midway, fill=white]  {F}}
      child {node (1_12) [word] {real}  edge from parent [white]}
      child {node (1_14) [word] {mechanic}  edge from parent node[midway, fill=white]  {C}}
      } edge from parent node[midway, fill=white]  {P/A}}
    child {node (1_10) [word] {check}  edge from parent node[midway, fill=white]{P}}
    } edge from parent node[midway, fill=white]  {H}}
  child {node (1_3) [word] {before}  edge from parent node[midway, fill=white]  {L}}
  child {node (1_4) [circle] {}
    {
    child {node (1_6) [word] {you}  edge from parent node[midway, fill=white]  {A}}
    child {node (1_7) [word] {leave}  edge from parent node[midway, fill=white]{P}}
    child {node (1_17) [word][text=red] {\textbf{IMP$_{Type-identifiable}$}}  edge from parent node[midway, fill=white]  {A}}
    } edge from parent node[midway, fill=white]  {H}}
  };
  \draw[dashed,->] (1_2) to node [midway, fill=white] {A} (1_6);
  \draw[bend right,->] (1_2) to[out=-20, in=180] node [midway, fill=white] {D} (1_12);
  \draw[red,very thick, ->] (1_9) to node [midway, fill=white] {A} (1_16);
\end{tikzpicture}
}
    \caption{In the predicted graph, a mismatched and a mislabelled implicit node are shown in red.}
    \label{fig:Evaluation example 1}
\end{subfigure}
\hfill
\begin{subfigure}[b]{0.8\textwidth}
\resizebox{10.5cm}{!}{
\begin{tikzpicture}
[->,level distance=1.35cm,
  level 1/.style={sibling distance=4cm},
  level 2/.style={sibling distance=25mm},
  level 3/.style={sibling distance=25mm},
  every circle node/.append style={fill=black},
  every node/.append style={text height=1ex,text depth=-1}]
  \tikzstyle{word} = [font=\rmfamily,color=black]
  \node (1_1) [circle] {}
  {
  child {node (1_2) [circle] {}
    {
    child {node (1_10) [word] {\textbf{IMP$_{Non-specific}$}}  edge from parent node[midway, fill=white]  {A}}
    child {node (1_4) [circle] {}
      {
      child {node (1_11) [word] {\textbf{IMP$_{Generic}$}}  edge from parent node[midway, fill=white]  {A}}
      child {node (1_12) [word][text=blue] {\textbf{IMP$_{Genre-based}$}}  edge from parent node[midway, fill=white]  {A}}
      child {node (1_7) [word] {The}  edge from parent node[midway, fill=white]  {F}}
      child {node (1_8) [word] {service}  edge from parent node[midway, fill=white] {P}}
      } edge from parent node[midway, fill=white]  {A}}
    child {node (1_5) [word] {is}  edge from parent node[midway, fill=white]  {F}}
    child {node (1_6) [word] {over-rated}  edge from parent node[midway, fill=white]  {S}}
    child {node (1_9) [word] {.}  edge from parent node[midway, fill=white]  {U}}
    } edge from parent node[midway, fill=white]  {H}}
  };
    \draw[blue,very thick, ->] (1_4) to node [midway, fill=white] {A} (1_12);
\end{tikzpicture}
}
\resizebox{10.5cm}{!}{
\begin{tikzpicture}[->,level distance=1.35cm,
  level 1/.style={sibling distance=4cm},
  level 2/.style={sibling distance=25mm},
  level 3/.style={sibling distance=25mm},
  every circle node/.append style={fill=black},
  every node/.append style={text height=1ex,text depth=0}]
  \tikzstyle{word} = [font=\rmfamily,color=black]
  \node (1_1) [circle] {}
  {
  child {node (1_2) [circle] {}
    {
    child {node (1_10) [word] {\textbf{IMP$_{Non-specific}$}}  edge from parent node[midway, fill=white]  {A}}
    child {node (1_4) [circle] {}
      {
      child {node (1_11) [word] {\textbf{IMP$_{Generic}$}}  edge from parent node[midway, fill=white]  {A}}
      child {node (1_7) [word] {The}  edge from parent node[midway, fill=white]  {F}}
      child {node (1_8) [word] {service}  edge from parent node[midway, fill=white] {P}}
      } edge from parent node[midway, fill=white]  {A}}
    child {node (1_5) [word] {is}  edge from parent node[midway, fill=white]  {F}}
    child {node (1_6) [word] {over-rated}  edge from parent node[midway, fill=white]  {S}}
    child {node (1_9) [word] {.}  edge from parent node[midway, fill=white]  {U}}
    } edge from parent node[midway, fill=white]  {H}}
  };
\end{tikzpicture}
}
    \caption{In the gold graph, one implicit node is marked in blue, indicating it is not matched in the predicted graph.}
    \label{fig:Evaluation example 2}
\end{subfigure}
\caption{Evaluation examples. Above: gold graphs. Below: predicted graphs.}
\end{figure*}

To compare a graph \(G_1\) with implicit units \(I_1\) to a graph \(G_2\) with implicit units \(I_2\) over the same sequence of terminals \(W=w_1, \dots, w_n\), for each implicit node \(i\), we identify its parent node \(p(i)\), denoting the set of terminals spanned by it as the yield \(y(p(i))\subseteq W\), and its category as the label \(\ell(p(i),i)\). Define the set of mutual implicit units between \(G_1\) and \(G_2\):
\begin{align*}
& M(G_1, I_1, G_2, I_2) = \\
& \left\{(i, j) \in I_1 \times I_2 \;\Bigg|\;
\begin{array}{ll}
y_1(p_1(i))=y_2(p_2(j)) \;\wedge \\
\ell_1(p_1(i),i)=\ell_2(p_2(j),j)
\end{array}
\right\}
\end{align*}
The F-score is the harmonic mean of labelled precision and recall, defined by dividing \(|M(G_1, I_1, G_2, I_2)|\) by \(|I_1|\) and  \(|I_2|\), respectively.

In labelled evaluation, it is worth noting that we require a full match of the two sets of labels/categories rather than just intersection, which suffices in standard UCCA evaluation (of non-implicit units).
That is, if there are two or more implicit nodes under one parent, it is only considered a correct match when both the numbers of implicit nodes and their labels are equal.
We also introduce unlabelled evaluation, which only requires that parents' spans match.
\begin{table*}[th]
\centering
\scalebox{0.6}{
\begin{tabular}{l|llllll|r}
 & Deictic & Generic & Genre-based & Type-identifiable & Non-specific & Iterated-set & \textit{Total Implicit Arguments} \\ \hline
\citet{cui-hershcovich-2020-refining} & 66 & 65 & 103 & 39 & 100 & 12 & \textit{385} \\
Our dataset & 107 & 86 & 147 & 6 & 36 & 9 & \textit{391}
\end{tabular}
}
\caption{Statistics of Revisited Implicit Corpus compared to the pilot annotation.}
\label{tab:Statistics of Revisited Dataset}
\end{table*}

For example, in Figure~\ref{fig:Evaluation example 1}, the reference graph has two implicit arguments. The first implicit unit's parent spans \{have, a, real, mechanic, check\} and the second's \{you, leave\}. In the predicted graph, two implicit units are predicted. The first implicit unit's parent spans \{a, real,  mechanic\} while the second one spans the same terminals as the reference graph. We can see that the spans of the first implicit unit do not match—the second matches but with the wrong label. Therefore, in labelled and unlabelled implicit unit evaluation, the precision, recall, and F1 scores are all $0$ and $0.5$, respectively.

In Figure~\ref{fig:Evaluation example 2}, the reference graph has three implicit units, labelled Non-specific, Generic and Genre-based. One spans \{The, service, is, overrated\} while two span \{The, service\}. Although the predicted graph has two implicit units with correct labels (Non-specific and Generic), it misses both implicit units under the second parent span, viz. the two sets merely intersect but are not equal. Therefore, the triplet precision, recall and F1 score for labelled evaluation are all $0.5$ and the triplet for unlabelled evaluation are $1$.

\subsection{Inter-annotator Agreement}
The two annotators separately reviewed and refined 15 out of the 116 passages (taken from the original test split). Using the evaluation metric proposed in \S\ref{sec:Evaluation}, the labelled and unlabelled F1 scores are 73.8\% and 91.3\% respectively (see Appendix~\ref{app:inter-annotator agreement}). Annotators have a Cohen's \(\kappa\) \citep{copenkappa} of 69.3\% on the six-type fine-grained classification of the implicit arguments whose parents' spans match.
For comparison, on the FiGref scheme, \citet{o2019bringing} report a Cohen's \(\kappa\) of 55.2\% on a 14-way classification and of 58.1\% on a four-way classification. \citet{gerber-chai-2010-beyond} proposed a relevant task to annotate implicit arguments for instances of nominal predicates in sentences, which has a Cohen's \(\kappa\) of 67\%. While we maintain a comprehensive fine-grained typology, we still see an improvement in agreement over other corpora.

\subsection{Statistics of Revisited Implicit Corpus}
Finally, one annotator reviewed and refined all 116 passages. The second reviewed their annotation after completion.
The full revisited dataset contains 393 passages, 3700 tokens and 5475 nodes. Table~\ref{tab:Statistics of Revisited Dataset} compares our revisited dataset to the unreviewed dataset of \citet{cui-hershcovich-2020-refining}. 

\begin{table*}[t]
\centering
	\begin{adjustbox}{width=375pt,margin=1pt,frame}
	\begin{tabular}{llll|l|lllll|l}
		\multicolumn{4}{c|}{\textbf{\small Before Transition}} &
		\multirow{2}{*}{\textbf{\small Transition}} & 
		\multicolumn{5}{c|}{\textbf{\small After Transition}} &\multirow{2}{*}{\textbf{\small Condition}} \\
		\textbf{\footnotesize Stack} & \textbf{\footnotesize Buffer} & 
		\textbf{\footnotesize Nodes} & \textbf{\footnotesize Edges} & & 
		\textbf{\footnotesize Stack} & \textbf{\footnotesize Buffer} & 
		\textbf{\footnotesize Nodes} & \textbf{\footnotesize Edges} & \textbf{\footnotesize Terminal?} &
		 \\ \hline
		$S$ & $x\;|\;B$ & $V$ & $E$ & \textsc{Shift} & $S\;|\;x$ & $B$ & $V$ & $E$ & $-$ & \\
		$S\;|\;x$ & $B$ & $V$ & $E$ & \textsc{Reduce} & $S$ & $B$ & $V$ & $E$ & $-$ & \\
		\textcolor{red}{$S\;|\;x$} & \textcolor{red}{$B$} & \textcolor{red}{$V$} & \textcolor{red}{$E$} & \textcolor{red}{\textsc{Node$_X$}} & \textcolor{red}{$S\;|\;x$} & \textcolor{red}{$y\;|\;B$} & \textcolor{red}{$V\cup\{y\}$} & \textcolor{red}{$E\;|\;(y,x)_{X}$} & \textcolor{red}{$-$} & $x \neq$ root \\
		\textcolor{blue}{$S\;|\;x$} & \textcolor{blue}{$B$} & \textcolor{blue}{$V$} & \textcolor{blue}{$E$} & \textcolor{blue}{\textsc{Node}} & \textcolor{blue}{$S\;|\;x$} & \textcolor{blue}{$y\;|\;B$} & \textcolor{blue}{$V\cup\{y\}$} & \textcolor{blue}{$E$} & \textcolor{blue}{$-$} & \\
		\textcolor{red}{$S\;|\;x$} & \textcolor{red}{$B$} & \textcolor{red}{$V$} & \textcolor{red}{$E$} & \textcolor{red}{\textsc{Implicit$_X$}}  & \textcolor{red}{$S\;|\;x$} & \textcolor{red}{$y\;|\;B$} & \textcolor{red}{$V\cup\{y\}$} & \textcolor{red}{$E\;|\;(x,y)_{X}^{\#}$} & \textcolor{red}{$-$} &  \\
		$S\;|\;y,x$ & $B$ & $V$ & $E$ & \textsc{Left-Edge$_X$} & $S\;|\;y,x$ & $B$ & $V$ & $E\;|\;(x,y)_{X}$ & $-$ &  \\
		$S\;|\;x,y$ & $B$ & $V$ & $E$ & \textsc{Right-Edge$_X$} & $S\;|\;x,y$ & $B$ & $V$ & $E\;|\;(x,y)_{X}$ & $-$ & $\left\{\begin{array}{l}x \notin w_{1: n}, \\ y \neq \text { root }, \\ y \wedge_{G} x\end{array}\right.$ \\
		$S\;|\;y,x$ & $B$ & $V$ & $E$ & \textsc{Left-Remote$_X$} & $S\;|\;y,x$ & $B$ & $V$ & $E\;|\;(x,y)_{X}^{*}$ & $-$ &  \\
		$S\;|\;x,y$ & $B$ & $V$ & $E$ & \textsc{Right-Remote$_X$} & $S\;|\;x,y$ & $B$ & $V$ & $E\;|\;(x,y)_{X}^{*}$ & $-$ &  \\
		$S\;|\;x,y$ & $B$ & $V$ & $E$ & \textsc{Swap} & $S\;|\;y$ & $x\;|\;B$ & $V$ & $E$ & $-$ & $\mathrm{i}(x)<\mathrm{i}(y)$ \\
		$[\mathrm{root}]$ & $\emptyset$ & $V$ & $E$ & \textsc{Finish} & $\emptyset$ & $\emptyset$ & $V$ & $E$ & $+$ & \\
	\end{tabular}
	\end{adjustbox}
	\caption{The transition sets of two implicit transition systems. Actions marked in \textcolor{red}{red} are for \textsc{Implicit-Eager}, \textcolor{blue}{blue} for \textsc{Implicit-Standard}. We write the \textbf{stack} with its top to the right and the \textbf{buffer} with its head to the left. $(\cdot, \cdot)_{X}$ denotes a X-labelled edge, $(\cdot, \cdot)_{X}^{*}$ a remote X-labelled edge, and $(\cdot, \cdot)_{X}^{\#}$ an X-labelled edge to an implicit node. $i(x)$ is a running index for the created nodes. The prospective child of the \textsc{Edge} action cannot have a primary parent. The newly generated node by \textcolor{red}{\textsc{Implicit$_X$}} action is prohibited from having any descendant. \textcolor{blue}{\textsc{Node}} generates a concept node on the buffer, but deos not produce an arc. This table is adapted from \citet{hershcovich2017a}.}
    \label{fig:transitions}	
\end{table*}
We see a major decline in Non-specific and Type-identifiable implicit arguments, because of the clearer definition of prominent elements and cases of agent nouns. Deictic, Generic and Genre-based increase their amount thanks to incorporating implicit arguments in State Scenes rather than only Process Scenes. The number of Iterated-set remains small due to the rareness of aspectual morphology and habitual/iterative constructions in English and in the corpus. However, it is still necessary to keep the category and separate out its instances rather than lump into another category or even ignore them. Since implicit arguments of such kind could be more common in morphologically rich languages, we want to keep a clean mapping of habitual/iterative constructions so as to facilitate the studies of implicit roles' diverse behaviours in languages other than English.

\section{Two Transition Systems for Parsing Implicit Arguments}\label{sec:parser}

We build the first neural parser that supports parsing implicit arguments dynamically in meaning representations, with two different transition systems. We design a transition-based parser, modelled upon \citet{nivre2003efficient}: a stack \(S=(\ldots,s_1,s_0)\) holds processed words. \(B=(b_0,b_1,\ldots)\) is a buffer containing tokens or nodes to be processed. \(V\) is a set of nodes, and \(E\) is a set of labelled edges. We denote \(s_{0}\) as the first element on \(S\) and \(b_{0}\) as the first element on \(B\). Given a sentence composed by a sequence of tokens \(t_{1}\), \(t_{2}\), ..., \(t_{n}\), the parser is initialized to have a Root node on \(S\), and all surface tokens in \(B\). The parser will at each step deterministically choose the most probable transition based on its current parsing state. Oracle action sequences are generated for training on gold-standard annotations.

We propose two transition systems, \textsc{Implicit-Eager} and \textsc{Implicit-Standard}, to deal with implicit arguments over the architecture of HIT-SCIR 2019 \cite{che-etal-2019-hit}, which ranked first in UCCA parsing in the MRP 2019 shared task \citep{oepen-etal-2019-mrp}. The transition system incorporates all nine transitions, namely, \textsc{Left-edge}, \textsc{Right-edge}, \textsc{Shift}, \textsc{Reduce}, \textsc{Node}, \textsc{Swap}, \textsc{Left-remote}, \textsc{Right-remote} and  \textsc{Finish}.

\textsc{Shift}, together with \textsc{Reduce}, are standard transitions.
\textsc{Shift} moves  \(b_{0}\) to \(S\), while 
\textsc{Reduce} pops \(s_{0}\) from \(S\) (when it should not be attached to any element in \(B\)).

Following transition-based constituent parsing, \textsc{Node}$_X$ creates a new non-terminal node \citep{sagae2005classifier}. Such node will be created on the buffer, as a parent of \(s_{0}\) with an \(X\)-labelled edge.

\textsc{Left-Edge}$_X$ and \textsc{Right-Edge}$_X$ add an \textsc{X}-labelled primary edge between the first two elements on \(S\). When the first element is the parent of the second element on \(S\), \textsc{Left-Edge}$_X$ is executed; in reverse, \textsc{Right-Edge}$_X$ will be chosen when the second element has the first element as its child. The left/right direction is the same as where the arc points to. \textsc{Left-Remote}$_X$ and \textsc{Right-Remote}$_X$ are similar to \textsc{Left-Edge}$_X$ and \textsc{Right-Edge}$_X$, yet these two transitions create remote edges, creating reentrencies. The \textsc{X}-labelled edge will be assigned a Remote attribute.

\textsc{Swap} deals with non-planar graphs (a generalisation of non-projective trees), in other words, discontinuous constituents. It pops the second node on \(S\) and adds it to the top of \(B\). \textsc{Finish} is the terminal transition, which pops the Root node and marks the transition state as terminal. 
\begin{table*}[th]
\centering
\scalebox{0.6}{
\begin{tabular}{l|llll|llllll|r}
 \textbf{Data} &\textbf{\# Sentences} &\textbf{\# Tokens} &\textbf{\# Nodes} &\textbf{\# Edges}  &\textbf{\# Deictic} &\textbf{\# Generic} &\textbf{\# Genre-based} &\textbf{\# Type-i} &\textbf{\# Non-s} &\textbf{\# Iterated-s} &\textbf{\# Implicit sum}\\ \hline
EWT Train & 2723 & 44751 & 59654 & 97561 & & & & & & & \\
EWT Dev & 554 & 5394 & 7534 & 11987 & & & & & & & \\
EWT Eval & 535 & 5381 & 7431 & 11907 & & & & & & & \\
\hline
Overall & & & &  & \textcolor{gray}{not refined} & \textcolor{gray}{not refined} & \textcolor{gray}{not refined} & \textcolor{gray}{not refined} & \textcolor{gray}{not refined} & \textcolor{gray}{not refined} & \textcolor{gray}{153} \\ 
\hline 
IMP Train & 285 & 2671 & 3936 & 6146 & 87 & 59 & 103 & 3 & 18 & 4 & 274 \\
IMP Dev   & 59  & 540  & 781  & 1217  & 11  & 15  & 19 & 1 & 10  & 0 & 56  \\
IMP Eval  & 49  & 489  & 709  & 1106  & 9 & 12 & 25 & 2  & 8  & 5 & 61  \\ \hline
Overall & & & &  & 107 & 86 & 147 & 6 & 36 & 9 & 391 \\
\end{tabular}
}
\caption{Statistics of train, dev and evaluation set in Original EWT and Revisited Implicit EWT. For each set, number of sentences, number of tokens, number of nodes, number of instances of 6 implicit categories and their sum are listed.}
\label{Statistics of Train Dev Eval dataset}
\end{table*}

\begin{table*}[t]
\centering
\scalebox{0.67}{
\begin{tabular}{l|lll|lll|cccccc}
 & \multicolumn{3}{c|}{Primary} & \multicolumn{3}{c|}{Remote} & \multicolumn{6}{c}{\textbf{Implicit}} \\
 & \textbf{LP} & \textbf{LR} & \multicolumn{1}{c|}{\textbf{LF}} & \textbf{LP} & \textbf{LR} & \multicolumn{1}{c|}{\textbf{LF}} & \textbf{LP} & \textbf{LR} & \textbf{LF} & \textbf{UP} & \textbf{UR} & \textbf{UF} \\
 \hline
Baseline & 0.495 & 0.467 & 0.480 & 0.538 & 0.304 & 0.389 & 1 & 0 & 0 & 1 & 0 & 0 \\
\hline
\textsc{Implicit-Eager} & 0.503 & 0.472 & 0.487 & 0.333 & 0.100 & 0.154 & \multirow{2}{1cm}{0.333 (7/21)} & \multirow{2}{1cm}{0.140 (7/50)} & 0.197 & \multirow{2}{1cm}{0.428 (9/21)} & \multirow{2}{1cm}{0.180 (9/50)} & 0.254 \\\\
\textsc{Implicit-Standard} & 0.474 & 0.431 & 0.451 & 0.438 & 0.280 & 0.341 & \multirow{2}{1cm}{\textbf{0.409} (9/22)} & \multirow{2}{1cm}{\textbf{0.180} (9/50)} & \textbf{0.250} & \multirow{2}{1cm}{\textbf{0.500} (11/22)} & \multirow{2}{1cm}{\textbf{0.220} (11/50)} & \textbf{0.306} \\\\
\end{tabular}
}
\caption{Experiment results on Revisited Implicit EWT in percents. For primary edges, remote edges, and implicit prediction, listed are Labelled Precision(LP), Labelled Recall (LR) and Labelled F-score (LF). In addition, Unlabelled precision (UL), Unlabelled Recall (UR) and Unlabelled F-score are also listed for implicit evaluation.}
\label{Experiment results on Revisited Implicit EWT}
\end{table*}
\section{Experiments}\label{sec:exp}
In \textsc{Implicit-Eager}, we introduce a new transition \textsc{Implicit}$_X$ adding an implicit node to the buffer and attaching it with a labelled edge in one step.
In \textsc{Implicit-Standard}, we simplify the existing \textsc{Node} transition to only create a node without attaching it, with the purpose of treating implicit units like primary ones and generating them dynamically. We elaborate their designs in Section~\ref{sec:Implicit-Eager} and Section~\ref{sec:Implicit-Standard}.
Table~\ref{fig:transitions} shows the transition set.

\subsection{\textsc{Implicit-Eager}}
\label{sec:Implicit-Eager}
Besides the nine transitions described above, \textsc{Implicit-Eager} introduces the \textsc{Implicit}$_X$ transition, which creates a new unit on \(B\) as the child of \(s_0\), with an \textsc{X}-labelled edge. The \textsc{Implicit} action is different from the \textsc{Node} action of \textsc{Implicit-Standard} in the sense that the integrally generated edge makes the new node a child of \(s_0\) rather than its parent, as in \textsc{Node}$_X$. Equally importantly, the new node is prohibited to have any child in contrast to the \textit{primary} nodes that the \textsc{Node}$_X$ action generates.

\subsection{\textsc{Implicit-Standard}}
\label{sec:Implicit-Standard}
\textsc{Implicit-Standard} adopts a more modular approach. Rather than complicating the transition systems, it treats primary non-terminal nodes and implicit nodes equally by simplifying the \textsc{Node}$_X$ action, making it generate a new unit on the buffer without attaching it with any (labelled) edge. We assume primary non-terminal nodes and implicit nodes are identical in essence, thus handling them without discrimination.  Whenever an ungenerated child or parent of \(s_0\) is found, \textsc{Node} is executed so that a concept node will be created on \(B\). This action does not cope with edge generation; the work is left to \textsc{Left-Edge}$_X$ or \textsc{Right-Edge}$_X$. In the oracle,  we can tell whether the node is primary or implicit by observing its relations. If the newly created node is the child of \(s_0\) and does not have any descendants, it is an implicit node; otherwise, a primary node.

\subsection{Data Preprocessing}
We convert UCCA XML data to MRP format using the open-source \texttt{mtool} software.\footnote{\url{https://github.com/cfmrp/mtool}} As the UCCA data provided in MRP 2019 shared task did not contain implicit information, HIT-SCIR 2019 is not designed to read this information in our dataset. We modify the parser to read node properties, and to convert UCCA data from and to MRP format. The updated version of \texttt{mtool} is available on GitHub.\footnote{\url{https://github.com/ruixiangcui/mtool}}

\subsection{Experimental Setup}

Our parsers use stack LSTM to stabilize gradient descent process and speed up training; we enrich contextual information by employing the pre-trained language model BERT as a feature input \citep{graves2013generating, devlin2018bert}. The model is implemented in AllenNLP \citep{Gardner2017AllenNLP}. We use the HIT-SCIR 2019 parser as the baseline for comparison. We keep the same hyperparameters as \citet{che-etal-2019-hit} except batch size, adjusted from 8 to 4 due to resource constraints. We do not tune hyperparameters on either the original or revisited dataset.\footnote{Hyperparameter settings are listed in Appendix~\ref{app:hyperparameters}.} 

We use the train, validation and evaluation split from \citet{hershcovich-etal-2019-semeval}, which was originally from UD EWT, with the ratio of 0.75, 0.125 and 0.125. The evaluation set has been validated as the gold standard.
Table~\ref{Statistics of Train Dev Eval dataset} shows detailed statistics of train, dev and eval set of both the original and revisited dataset, on which we trained the baseline parser, \textsc{Implicit-Eager} and \textsc{Implicit-Standard}.\footnote{See training details in Appendix~\ref{app:training details}.}

\section{Results}\label{sec:results}

Table~\ref{Experiment results on Revisited Implicit EWT} presents experimental results on Revisited Implicit EWT by three parsers, the baseline HIT-SCIR 2019 parser, \textsc{Implicit-Eager} and \textsc{Implicit-Standard} on Revisited Implicit EWT. Regarding performance on the dataset, the baseline is not able to predict implicit argument as expected. However, both \textsc{Implicit-Eager} and \textsc{Implicit-Standard} managed to predict implicit arguments.  

Based on the evaluation method mentioned in section~\ref{sec:Evaluation}, \textsc{Implicit-Eager}'s labelled precision and labelled recall on Revisited Implicit EWT are 0.333 and 0.14; the unlabelled precision and unlabelled recall are 0.428 and 0.18. For the primary edge and remote edge evaluation, noticeably, \textsc{Implicit-Eager} also outperforms the baseline on primary edges by 0.007 in F-score on the revisited dataset. 

Even though \textsc{Implicit-Standard} has the worse results in terms of primary parsing, it gains boosted performance on all targets in implicit evaluation. Its unlabelled implicit precision, recall and F-score are 0.5, 0.22 and 0.306, defeating \textsc{Implicit-Eager} by 0.072, 0.04 and 0.052, respectively.

For labelled evaluation, \textsc{Implicit-Standard} surpasses \textsc{Implicit-Eager} even further; the labelled precion, recall and F-score are 0.419, 0.180 and 0.250, respectively exceeding \textsc{Implicit-Eager} by 0.076, 0.04 and 0.053.

Table~\ref{Experiment results on Original EWT} presents the three parsers' performances on Original EWT. The baseline produced better results on primary edges and remote edges on Original EWT. 

The reason why \textsc{Implicit-Standard} outperforms \textsc{Implicit-Eager} on implicit evaluation but decrease in accuracy on primary evaluation might be attributed to its equal treatment of primary nodes and implicit nodes.

\begin{table}[t]
\scalebox{0.66}{
\begin{tabular}{l|lll|lll}
 & \multicolumn{3}{c|}{Primary} & \multicolumn{3}{c}{Remote} \\
 & \textbf{LP} & \textbf{LR} & \multicolumn{1}{c|}{\textbf{LF}} & \textbf{LP} & \textbf{LR} & \multicolumn{1}{c}{\textbf{LF}} \\
 \hline
Baseline & 0.710 & 0.701 & 0.706 & 0.547 & 0.365 & 0.438 \\
\hline
\textsc{Implicit-Eager} & 0.675 & 0.597 & 0.634 & 0.527 & 0.344 & 0.416 \\
\textsc{Implicit-Standard} & 0.656 & 0.571 & 0.610 & 0.458 & 0.225 & 0.302 \\
\end{tabular}
}
\caption{Experiment results on Original EWT in percents. As there is no implicit argument in the dataset, only performances on primary edges and remote edges are listed.}
\label{Experiment results on Original EWT}
\end{table}

\section{Discussion}\label{sec:discussion}
 As is indicated in Table~\ref{Experiment results on Revisited Implicit EWT}, \textsc{Implicit-Eager} and \textsc{Implicit-Standard} successfully predicted respectively seven and nine implicit arguments with the correct fine-grained implicit labels. In the unlabelled evaluation, nine and 11 implicit arguments were predicted each.
 
 Table~\ref{tab:Confusion matrix on the evaluation set of Revisited Implicit EWT} shows the confusion matrix of the performances of \textsc{Implicit-Eager} and \textsc{Implicit-Standard} on the evaluation set of Revisited Implicit EWT.  Both parsers have predicted roughly the same amount of implicit arguments, 22 and 21, respectively. 
 
Noticeably, \textsc{Implicit-Eager} has emitted 12 Deictic implicit arguments, accounting for 57.1\% of all predictions, 14 and 66.7\% if including partial matches as Deictic \& Generic and Deictic \& Genre-based. While \textsc{Implicit-Standard} has a more uniform distribution over prediction categories. Deictic, Generic and Genre-based are the most predicted ones. Moreover, one Non-specific implicit argument, despite being wrongly labelled, is also predicted by \textsc{Implicit-Standard}, while \textsc{Implicit-Eager} has never predicted labels other than Deictic, Generic and Genre-based, which have significantly more instances in the training set. 

\begin{table}[th]
\resizebox{\columnwidth}{!}{%
\begin{tabular}{l|cccccccccccc|r}
\textbf{\textsc{Implicit-Eager}} & \rotatebox{90}{UNMATCHED} & \rotatebox{90}{Non-specific} & \rotatebox{90}{Non-specific \& Generic} & \rotatebox{90}{Non-specific \& Type-identifiable} & \rotatebox{90}{Deictic} & \rotatebox{90}{Deictic \& Iterated-set} & \rotatebox{90}{Generic} & \rotatebox{90}{Generic \& Genre-based} & \rotatebox{90}{Genre-based} & \rotatebox{90}{Iterated-set} & \rotatebox{90}{Type-identifiable} & \rotatebox{90}{P} & \rotatebox{90}{\textit{Total}}\\
\hline
UNMATCHED & 0 & 5 & 1 & 1 & 4 & 1 & 1 & 7 & 14 & 4 & 1 & 2 & \textit{41}\\
Deictic & 6 & 1 & 0 & 0 & \textbf{4} & 0 & 0 & 0 & 1 & 0 & 0 & 0 & \textit{12}\\
Deictic \& Generic & 1 & 0 & 0 & 0 & 0 & 0 & 0 & 0 & 0 & 0 & 0 & 0 & \textit{1}\\
Deictic \& Genre-based & 1 & 0 & 0 & 0 & 0 & 0 & 0 & 0 & 0 & 0 & 0 & 0 & \textit{1}\\
Generic \& Genre-based & 1 & 0 & 0 & 0 & 0 & 0 & 0 & \textbf{3} & 0 & 0 & 0 & 0 & \textit{4}\\
Genre-based & 3 & 0 & 0 & 0 & 0 & 0 & 0 & 0 & 0 & 0 & 0 & 0 & \textit{3}\\
\hline \hline
\textbf{\textsc{Implicit-Standard}} &  &  &  &  &  &  &  &  &  &  &  &  \\
\hline
UNMATCHED & 0 & 6 & 1 & 1 & 4 & 1 & 1 & 6 & 12 & 4 & 1 & 2 & \textit{39}\\
Non-specific & 1 & 0 & 0 & 0 & 0 & 0 & 0 & 0 & 0 & 0 & 0 & 0 & \textit{1}\\
Deictic & 2 & 0 & 0 & 0 & \textbf{4} & 0 & 0 & 0 & 0 & 0 & 0 & 0 & \textit{6}\\
Deictic \& Genre-based & 0 & 0 & 0 & 0 & 0 & 0 & 0 & 1 & 0 & 0 & 0 & 0 & \textit{1}\\
Generic & 2 & 0 & 0 & 0 & 0 & 0 & 0 & 0 & 0 & 0 & 0 & 0 & \textit{2}\\
Generic \& Genre-based & 3 & 0 & 0 & 0 & 0 & 0 & 0 & \textbf{3} & 0 & 0 & 0 & 0 & \textit{6}\\
Genre-based & 3 & 0 & 0 & 0 & 0 & 0 & 0 & 0 & \textbf{2} & 0 & 0 & 0 & \textit{5}\\
Genre-based \& P & 0 & 0 & 0 & 0 & 0 & 0 & 0 & 0 & 1 & 0 & 0 & 0 & \textit{1}
\end{tabular}%
}
\caption{Confusion matrix on the Revisited Implicit EWT evaluation set: The column is the predicted labels while the row is the actual labels. Noticeably, the parsers are able to predict implicit elements of other categories in theory, such as Process (P). If not clarified otherwise, the fine-grained implicit categories are Participant by default.}
\label{tab:Confusion matrix on the evaluation set of Revisited Implicit EWT}
\end{table}

Both implicit parsers have predicted correctly four Deictic and three Generic \& Genre-based implicit arguments. Besides, \textsc{Implicit-Standard} managed to predict two more correct Genre-based while \textsc{Implicit-Eager} has never predicted successfully stand-alone Genre-based implicit argument. \textsc{Implicit-Standard} has a remarkably higher labelled precision of 66.7\% in Deictic than \textsc{Implicit-Eager} of 33.3\%, while the latter has a higher precision of 75\% in Generic \& Genre-based than \textsc{Implicit-Standard} of 50\%. Lamentably, neither of the implicit parsers has emitted prediction for Type-identifiable nor Iterated-set. It is necessarily expected as both categories have less than five instances in the training set.

Although this paper focuses on fine-grained implicit Participants, there are already some implicit arguments of other categories annotated in the foundational layer, especially Process and Center. Interestingly, in the sentence ``Fresh and excellent quality,'' \textsc{Implicit-Standard} generated two implicit arguments, Genre-based and Process in the Scene ``Fresh.'' It means not only that it infers the Scene misses a Genre-based Participant, but also a Process, i.e., the main relation that evolves in time.

\section{Related Work}\label{sec:related}

Parsing implicit argument was introduced into NLP by \citet{ruppenhofer-etal-2009-semeval,gerber-chai-2010-beyond,gerber-chai-2012-semantic}, but has been coarse-grained and annotated within Nombank \citep{meyers-etal-2004-nombank}, limited to ten nominal predicates.

\citet{bender-etal-2011-parser} identified ten relevant linguistic phenomena, ran several parsers and associated their output with target dependencies.
\citet{roth-frank-2015-inducing} used a rule-based method for identifying implicit arguments, which depends on semantic role labelling and coreference resolution.
Similarly, \citet{silberer-frank-2012-casting, chiarcos-schenk-2015-memory, schenk-chiarcos-2016-unsupervised} proposed parsing methods for SemEval 2010 data, but they are only able to parse implicit arguments on a coarse level as well.
\citet{cheng-erk-2018-implicit} built a narrative cloze model and evaluated it on \citet{gerber-chai-2010-beyond}'s dataset.

\section{Conclusion}\label{sec:conclusion}
Implicit arguments are pervasive in the text but have not been well studied from a general perspective in NLP. In this work, we revisited a recently proposed fine-grained implicit argument typology by addressing its current deficiencies. We annotated a corpus based on the revised guidelines and designed an evaluation metric for measuring implicit argument parsing performance, demonstrating the annotation's reliability with a superior inter-annotator agreement comparing to other fine-grained implicit argument studies. The dataset will be available to facilitate relevant research.

We introduced the first semantic parser, with two different transition systems, that can handle and predict implicit nodes dynamically, and label them with promising accuracy as part of the meaning representations. We evaluated it on the new dataset and found that some types of implicit arguments are harder to parse than others and that a simpler transition system performs better on parsing implicit arguments at the cost of primary parsing. The fine-grained implicit argument task is challenging and calls for further research. 

In future work, we plan to create a large resource of implicit arguments by automatically extracting them from various linguistic constructions in unlabelled text, use it for pre-training of language models, and evaluate them on our and other datasets to gain more insights into this linguistic phenomenon. A post-processing baseline to find implicit arguments after parsing the whole graph would also be interesting for future investigation.

\section{Acknowledgments}
The authors thank Miryam de Lhoneux and the anonymous reviewers for their helpful feedback.

\bibliography{references,anthology}
\bibliographystyle{acl_natbib}

\clearpage
\onecolumn
\appendix
\section*{Appendices}
\label{sec:appendix}

\section{Inter-annotator Confusion Matrix}
\label{app:inter-annotator agreement}
Tabel \ref{The evaluation set before reaching an agreement} shows the confusion matrix for measuring inter-annotator agreement on the evaluation set. The unlabelled F-score is 91.3\%, and the labelled F1-score is 73.8\%. The Cohen's \(\kappa\) between two annotators is 69.3\%.

\begin{table*}[ht]
\resizebox{\textwidth}{!}{%
\begin{tabular}{l|cccccccccccc}
 & UNMATCHED & Nonspecific & Nonspecific|Generic & Nonspecific|Type-identifiable & Deictic & Deictic|Iterated-set & Generic & Generic|Genre-based & Genre-based & Iterated/repeated/set & Type-identifiable & P \\
 \hline
UNMATCHED & 0 & 1 & 0 & 0 & 0 & 0 & 1 & 0 & 1 & 1 & 1 & 0 \\
Nonspecific & 0 & 4 & 0 & 0 & 0 & 0 & 0 & 0 & 4 & 0 & 0 & 0 \\
Nonspecific|Deictic & 0 & 0 & 0 & 0 & 0 & 1 & 0 & 0 & 0 & 0 & 0 & 0 \\
Nonspecific|Generic & 0 & 0 & 1 & 0 & 0 & 0 & 0 & 0 & 0 & 0 & 0 & 0 \\
Nonspecific|Genre-based & 0 & 0 & 0 & 0 & 0 & 0 & 0 & 1 & 0 & 0 & 0 & 0 \\
Nonspecific|Type-identifiable & 0 & 0 & 0 & 1 & 0 & 0 & 0 & 0 & 0 & 0 & 0 & 0 \\
Deictic & 2 & 0 & 0 & 0 & 8 & 0 & 0 & 0 & 0 & 0 & 0 & 0 \\
Deictic|Genre-based & 0 & 0 & 0 & 0 & 0 & 0 & 0 & 1 & 0 & 0 & 0 & 0 \\
Generic|Genre-based & 0 & 0 & 0 & 0 & 0 & 0 & 0 & 8 & 0 & 0 & 0 & 0 \\
Genre-based & 0 & 2 & 0 & 0 & 0 & 0 & 0 & 0 & 11 & 0 & 0 & 0 \\
Type-identifiable & 1 & 0 & 0 & 0 & 0 & 0 & 0 & 0 & 0 & 0 & 0 & 0 \\
Iterated-set & 1 & 0 & 0 & 0 & 0 & 0 & 0 & 0 & 0 & 3 & 0 & 0 \\
P & 0 & 0 & 0 & 0 & 0 & 0 & 0 & 0 & 0 & 0 & 0 & 2
\end{tabular}
}
\caption{Confusion matrix of the evaluation set for measuring inter-annotator agreement}
\label{The evaluation set before reaching an agreement}
\end{table*}

\section{Hyperparameter Settings}
\label{app:hyperparameters}
Our parsers use stack LSTM to stabilize gradient descent process and speed up training; we enrich contextual information by employing the pre-trained language model BERT as a feature input. We keep the same hyperparameter setting for all three parsers, the baseline parser HIT-SCIR 2019, \textsc{Implicit-Eager} and \textsc{Implicit-Standard}. The setting is shown as the Table \ref{tab:hyperparameters}.
\begin{table}[ht]
\centering
\begin{tabular}{l|r}
\multicolumn{1}{l|}{\textbf{Hyperparameter}}                                     & \multicolumn{1}{l}{\textbf{Value}} \\ \hline
Hidden dimension & 20 \\ 
    Action dimension & 50 \\
Optimizer & Adam \\
$\beta_{1}, \beta_{2}$ & 0.9, 0.99 \\
Dropout & 0.5 \\
Layer dropout & 0.2 \\
Recurrent dropout & 0.2 \\
Input dropout & 0.2 \\
Batch size & 4 \\
Epochs & 50 \\
Base learning rate & $1 \times 10^{-3}$ \\
BERT learning rate & $5 \times 10^{-5}$ \\
Gradient clipping & 5.0 \\
Gradient norm & 5.0 \\ \hline
Learning rate scheduler & slanted triangular \\
Gradual Unfreezing & True \\
Cut Frac & 0.1 \\
Ratio& 32 \\
\end{tabular}
\caption{Implicit Parser hyperparameters.}
\label{tab:hyperparameters}
\end{table}

\section{Training Details}
\label{app:training details}
As Table \ref{tab:training_details} shows, the training time is 2 days 22 hours for the baseline on Original UCCA EWT (50 epochs). Best epoch is 3rd; 3 hours for the baseline on Revisted Implicit EWT (30 epochs). Best epoch is 22nd; 1 day 8 hours and 1 day 19 hours for \textsc{Implicit-Eager} and \textsc{Implicit-Standard} on Original UCCA EWT (10 epochs, 13 epochs), respectively, with the best epoch being the 3rd and 2nd; And finally, 6 hours and 8 hours for \textsc{Implicit-Eager} and \textsc{Implicit-Standard} on Revisited Implicit EWT (50 epochs). The best epoch is 21st and 37th, respectively. One can see that all parsers achieved the best performance at an early stage on Original UCCA EWT. However, both implicit parsers took longer time to train on Original UCCA EWT than the baseline.
 
\begin{table*}[ht]
    \begin{tabular}{l|lll|lll}
     & \multicolumn{3}{c|}{Original Full UCCA EWT} & \multicolumn{3}{c}{Revisited Implicit Dataset} \\
     & \small Training time & \small \# Epochs & \small \# Best Epoch & \small Training time & \small \# Epochs & \small \# Best Epoch \\ \hline
    Baseline & 2 days 22 h & 50 & 3 & 3 h & 30 & 22 \\ 
    \textsc{Implicit-Eager} & 1 day 8 h & 10 & 3 & 6 h & 50 & 21 \\
    \textsc{Implicit-Standard} & 1 day 19 h & 13& 2 & 8 h & 50 & 37 \\
    \end{tabular}
    \caption{Training details of the baseline, \textsc{Implicit-Eager} and \textsc{Implicit-Standard} on orignial UCCA EWT and Revisited Implicit EWT., including training times, the number of best epoch and total epochs.}
    \label{tab:training_details}
\end{table*}

\end{document}